\relax
\documentclass[letterpaper]{article} %
\usepackage{graphicx} %

\usepackage[style=nature]{biblatex}
\title{Controlling for Biasing Signals in Images for Prognostic Models: Survival Predictions for Lung Cancer with Deep Learning}
\author{W.A.C. van Amsterdam and M.J.C. Eijkemans\\
    University Medical Center Utrecht\\
    Heidelberglaan 100, 3584CX, Utrecht\\
    Utrecht, The Netherlands\\
    w.a.c.vanamsterdam@umcutrecht.nl}
 
\begin{document}

\maketitle

\begin{abstract}
Deep learning has shown remarkable results for image analysis and is expected to aid individual treatment decisions in health care. 
To achieve this, deep learning methods need to be promoted from the level of mere associations to being able to answer causal questions. 
We present a scenario with real-world medical images (CT-scans of lung cancers) and simulated outcome data. 
Through the sampling scheme, the images contain two distinct factors of variation that represent a collider and a prognostic factor. 
We show that when this collider can be quantified, unbiased individual prognosis predictions are attainable with deep learning. 
This is achieved by (1) setting a dual task for the network to predict both the outcome and the collider and 
(2) enforcing independence of the activation distributions of the last layer with ordinary least squares. 
Our method provides an example of combining deep learning and structural causal models for unbiased individual prognosis predictions.\end{abstract}

\subsection{Introduction}

Deep learning has many possible applications in health care, especially 
for tasks including unstructured data such as medical images. 
Convolutional neural networks (CNN) are especially attractive for supervised tasks as they can be 
optimized end-to-end, 
and may detect patterns in the images that are relevant to the prediction task, 
but may be unknown to medical professionals.
A downside is that the induced representations of the network are hidden and 
not readily interpretable, though visualization techniques are being developed.
A much sought after \textit{holy grail} of artificial intelligence is to attain personalized
treatment decisions through individual prognosis prediction and individual treatment
effect estimation. 
This is a \textit{causal} question, and answering it requires techniques
from causal inference \cite{pearl-causality}. 
A pivotal result from causal inference is that when the structural causal model underlying the data-generating
mechanism is known, identifiability and estimands of causal queries can be deduced from the \textit{Directed Acyclic Graph} (DAG), 
which represents the known causal relationships between relevant variables. 

\subsubsection{Images and DAGs}
The connection between medical images and a DAG is not always straightforward to see. 
Fundamentally, patient outcomes are driven
by biological processes, and images may contain (more or less noisy) views of these processes. 
For example, a particularly aggressive lung tumor may grow very large, as can be seen on a CT-scan,
and this biological behavior leads to a worse expected survival. 
These biological processes
can be seen as underlying \textit{factors of variation} that \textit{cause} the image in the
sense of structural causal models. 
Conversely, information derived from medical images
is often used to make treatment decisions. Here, the image is a causal factor for 
treatment selection. 
In general, when a deep learning network is used to predict a certain clinical outcome, 
it will use all information in an image that is $statistically$ 
associated with that outcome. Predicting an outcome with deep learning based on an image
can be seen as conditioning on (noisy views of) causal factors of variations of these images.
Medical images, especially images from large body parts
such as a chest CT-scan in the case of lung cancer, may contain many different factors of variation
that can have different 'roles' in the DAG. 
Notably when a specific factor of variation represents a \textit{collider} in the DAG, conditioning on the image by using a deep learning model may introduce bias in the estimation of causal effects. 

\subsubsection{Problem setup}
We describe a fictional but realistic
clinical scenario where the following conditions hold: 
(1) There exists a clinical need for outcome prediction
(2) This outcome partly depends on treatment, and an unbiased estimate of the treatment
effect is required
(3) The DAG describing the data-generating process is assumed to be known
(4) An image is hypothesized to contain important information for the task in (1), 
however, one of the factors of variation causing the image represents a collider in the DAG. 
Conditioning on this collider will lead to a biased estimate of (2)
(5) The collider can be measured from the image
(6) Deep learning is used to optimally predict (1).
We stress that this poses a conflicting problem: 'simply' using deep learning to predict the outcome based on the image 
may lead to a low prediction error of the outcome in the observational setting, 
but it will lead to bias in the estimated effect of treatment, as it conditions on a collider.
This means that it does not give accurate predictions in the setting where we \textit{intervene} on treatment, 
the predictions are only valid when treatment was assigned according to the regular clinical care through which the data where gathered.
On the other hand, ignoring the image all together will lead to worse prediction error. 
Our contribution is that we show that by utilizing a multi-task prediction scheme for both the outcome and the collider, 
accompanied by an additional loss term to induce independence between final layer activations, 
we can satisfy both (1) the supervised prediction task and (2) attain an unbiased estimate of the treatment effect. 
For clarity in notation, we will reserve the term \textit{prediction error} for performance on the supervised prediction task 
(e.g. accuracy of predicted survival time). 
With \textit{bias} we will refer to difference between the expectation of the estimated treatment effect and the data-generating mechanism.

\subsection{Methods}

\subsubsection{Clinical case}

The proposed clinical case concerns the treatment of lung cancer. 
Optimal treatment selection for lung cancer patients is a challenging problem: 
depending on the disease stage, patients receive (combinations of)
chemotherapy, surgery, or recently, immunotherapy or targeted therapy \cite{nccn}. 
Some patients will be cured, while others only endure invalidating side-effects.
Personalized treatment decisions may be aided by estimating the individual prognosis of a patient for the different modes of treatment that are available.
Medical scans provide important information for diagnosing and staging lung cancer, 
but may also provide this prognostic information.
Deep learning is particularly attractive to analyse these scans, 
as these models may discover new,
previously unused prognostic factors or treatment effect modifiers.

\subsubsection{Data generating mechanism}
\begin{figure}
  \centering
  \includegraphics[width=1.0\textwidth]{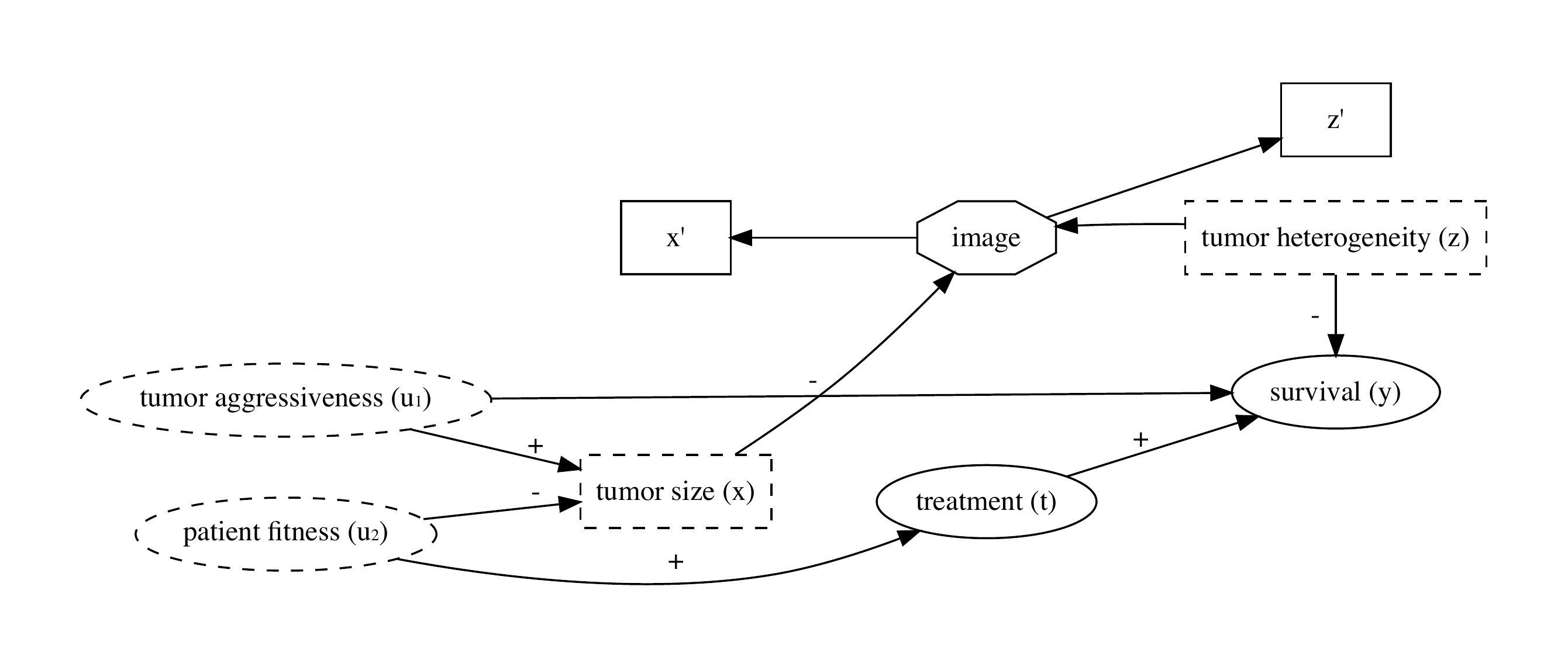}
  \caption{Directed Acyclic Graph describing the data-generating mechanism for the simulations. 
          Signs indicate positive or negative associations. Rectangle shaped variables are image variables,
          dashed variables are unobserved. $x, z$ represent biological processes, 
          causing the outcome and image patterns, 
          $x', z'$ are noisy views of these variables that are measurable from the image.}
  \label{fig:dag}
\end{figure}

In our experiments we use a real-world data set of lung cancer scans from the 
Lung Image Database Consortium image collection (LIDC \cite{lidc}). 
These 1018 scans each contain lung nodules ($N = 2609$) suspected of lung cancer.
Up to 4 radiologists segmented the nodules on each consecutive image slice,
resulting in accessible measurements of the nodule sizes. 
A CT-scan measures radiodensity, and tissues may exhibit different density-patterns.
Heterogeniety in radiodensity is known to be associated with worse biologic aggressiveness
and survival \cite{heterogeneity}.
We used nodule size and the variance of radiodensity in a simulated binary treatment and real-valued outcome model.
Figure \ref{fig:dag} and Table \ref{tabsims} illustrate the following hypothetical narrative:

There exist two possible treatments for lung cancer: $t \in \{0,1\}$,
where $t=1$ is deemed more aggressive and also more effective. 
An unobserved noise variable $u_2$ influences treatment allocation: 
people who appear to be in better overall health, as per subjective judgement of the physician,
will have a higher probability of being treated with $t=1$. 
At the same time they generally have a better functioning immune system. 
The immune system combats the lung cancer, leading to a lower tumor size ($x$).
Another unobserved noise variable $u_1$ represents the tumor biologic aggressiveness. 
High aggressiveness leads to a bigger tumor and negatively impacts the overall survival. 
We emphasize that the tumor size ($x$) is a pre-treatment collider according to this causal graph. 
A third noise variable, variance of radiodensity ($z$), is a prognostic factor unrelated to the treatment, but related to the outcome. 
Tumors with high variance in radiodensity are more aggressive and lead to reduced survival.

This situation leads to a conundrum. 
As can be seen from the DAG, the marginal average treatment effect is identified
by $ATE = E[p(y|t=1) - p(y|t=0)]$. The conditional treatment effect 
is not identified when conditioning the entire image, which is a descendant of both $x$ and $z$.
Conditioning on $x'$ (the tumor size as measured in the image), 
corresponds to partly conditioning the collider $x$.
This will induce dependence between $u_1$ and $u_2$,
thereby opening a confounding path from $t$ to $y$ and violating of the backdoor criterion \cite{pearl-causality}.
Using a convolutional neural network to predict $y$ without regard for the biasing effect 
of conditioning on the collider will lead to a biased estimate of the treatment effect. 
Disentangling the factors of variation in the image to only utilize image information that not related to the collider
would enable an unbiased estimate of the $ATE$, which is the goal of this study.

\begin{table}[]
  \centering
  \setlength\tabcolsep{1.5pt}
  \begin{tabular}{lll}
    & \textbf{variable} & \textbf{variable model}    \\ \hline
    $u_1$             & aggressiveness & $N(0,0.7071)$               \\                  
    $u_2$             & fitness        & $N(0,0.7071)$               \\
    $z$               & heterogeneity  & $N(0, 1)$                \\
    $x$               & size           & $N(u_1 - u_2; 0.05)$        \\
    $t$               & treatment      & $Bern(logit(N(1.828 u_2 - 0.5, 0.25)))$ \\
    $y$               & survival       & $N(t - z - 2 u_1 - 0.5; 0.05)$  \\ \hline
  \end{tabular}
  \caption{Parameters for sampling images and modeling outcome data. For each observation $i$, 
           an image was drawn from the total pool of images with the closest $x_i$ and $z_i$.
           This ensures the required association between factors of variation in the image and the simulated outcome data}
  \label{tabsims}
\end{table}  

\begin{figure*}
  \centering

  \includegraphics[width=1.0\textwidth]{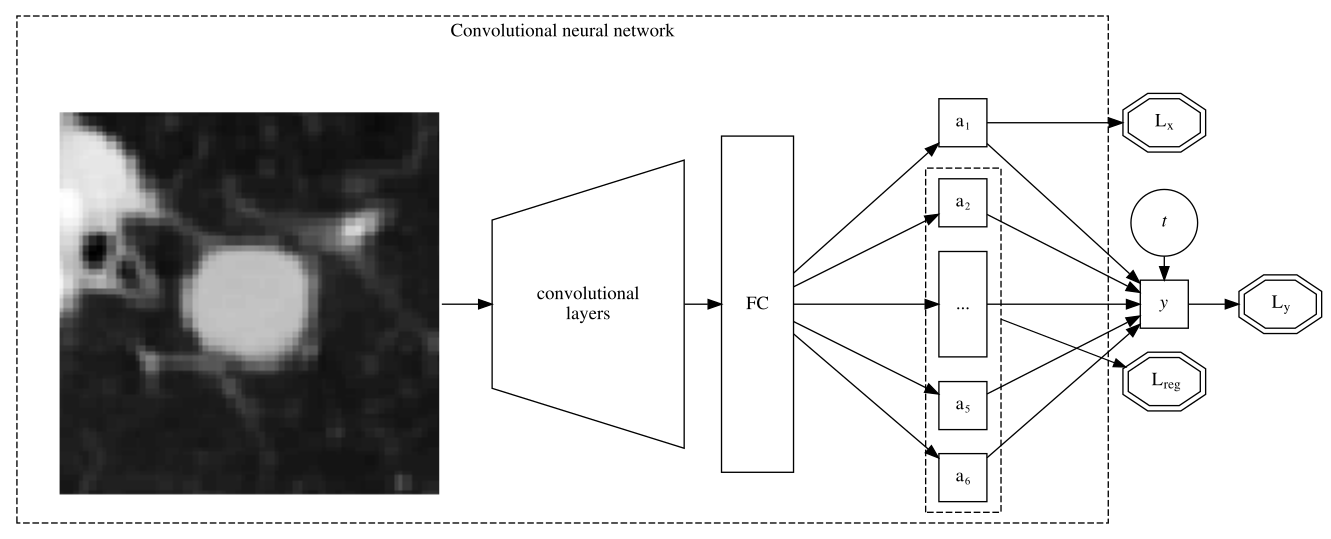}
  \caption{
    Schematic overview of convolutional neural network architecture. 
    The network receives two inputs: an image and the treatment indicator ($t$).
    Loss functions are depicted in double octagons. 
    The last layer activations are used to separate factors of variation 
    in the image. $a_1$ is trained to approximate $x$. 
    The rest of the last layer activations are constrained to be linearly 
    independent from $x$ through $L_{reg}$.
    The total loss is $L = L_y + L_x + L_{reg}$. 
    FC: fully-connected layers}
  \label{fig:cnn}
\end{figure*}
 
\subsubsection{Modeling}

Our method revolves around two central notions:
1. Utilizing the resemblance of the final layer of a CNN with linear regression
2. Separating the contributions of different factors of variation during training, to enable ablation after model convergence.
For each patient we have three observed quantities: $x_i, y_i \in {\rm I\!R}$ and $t_i \in \{0,1\}$. 
Following standard practice for predicting a continuous real outcome with deep learning, 
the last layer of the CNN resembles linear regression where 
$\hat{y} = \beta_0 + \beta_t t + \sum_{j=1}^{N_k}{\beta_{j}^k a_j^k}$, 
with $a_j^k$ the $N_k$ activations of the final layer of a $k$-layer CNN, 
$t$ the binary treatment indicator and $\beta_0$ an overall intercept. 
Indices for samples are omitted for clarity. 
Note that $\beta_t$ is the estimated average treatment effect ($ATE$).
The standard minibatch mean squared error is used for $y$: 
$$L_y = \frac{1}{m}\sum_{i=1}^m{(\hat{y} - y)^2}$$
where $m$ the minibatch size. 
During training we force a single activation of the last layer to be the best possible approximation of the collider: $a_1^k \approx x$.
At the same time we restrain the other last layer activations $\{a_j^k, j>1\}$ to be linearly independent of $x$.
Note that this is a light constraint based on the prior knowledge represented in the DAG, 
namely that $x$ is a scalar and $x$ and $z$ are independent.
We argue that after model convergence, we can fix all CNN parameters and do a single ordinary least squares on $\{a_j^k \cup t, j > 1\}$
to get a valid estimate of the treatment effect with $\beta_t$, as this 
separates out the biasing effect of the collider, which is 'contained' in $a_1^k$. 
To attain this, we add a dual target for the collider $x$:

$$L_x = \frac{1}{m}\sum_{i=1}^m{(a_1^k - x)^2}$$

This encourages the model to have a single activation in the last layer to 
approximate the collider $x$. 
Both losses are synergistic,
as predicting $x$ will improve $L_y$ since $x$ and $y$ are statistically associated.
At each training step, a prediction $\hat{x}^{reg} := \beta_0^{reg}+\sum_{j=2}^{N_k}{\beta_j^{reg}a_j^k}$ is made 
by fitting $\mathbf{\beta}^{reg}$ with ordinary least squares regression on the minibatch. 
The $MSE$ of this regression represents how well $x$ can be predicted from a linear combination of
the last layer activations $\{a_j^k, j>1\}$.
This is compared to the $MSE$ of predicting $x_i$ with $\bar{x}$, the 
mean of $x$ of that minibatch. When predicting $x$ from $\{a_j^k, j>1\}$ is no better 
than using the mean of $x$, 
we pose that these activations are sufficiently 
independent from $x$. 
Whenever the regularizing $MSE$ is lower than this mean approximation, this difference is added to the total loss.

$$L_{reg} := max(0, MSE(\bar{x},x) - MSE(\hat{x}^{reg}, x))$$

The total loss is the direct sum of these losses.

$$L = L_y + L_x + L_{reg}$$

Training was continued until convergence or overfitting, as assessed by an increase in total loss 
on an independently simulated validation set with different images than in the training set. 
After convergence, all CNN parameters were fixed and the final layer
activations calculated for each image. 
A linear regression of $y$ was fitted on $\{a_j, t; 1<j\leq N_k\}$ 
resulting in a final model, dubbed 'CausalNet'.

\subsection{Experiments}
Observations were generated by sampling noise variables from the appropriate distributions,
and dependent variables according to the structural causal model in Table \ref{tabsims}. For 
each patient $i$ with $x_i, z_i$, an image was drawn from the total pool of images 
with the closest measured $x, z$. 
We simulated 3000 training samples and 1000 validation samples.
Images from the training set where from different patients than from the validation set.
Square slices of 7x7cm surrounding the nodules were extracted from the CT-scan and 
resampled to isotropic 0.7mm spacing. Pixel intensities were normalized 
to unit scale using a global mean and variance. The images were cropped randomly to 51x51 pixels 
during training, center crops of the same size were used for validation.
Also random vertical and horizontal mirroring was used as data augmentation during training. 
A simple CNN architecture with 4 layers of 3x3 convolutions with 16 feature channels, each followed by ReLU non-linearity 
and 2x2 max-pooling was used. These basic image features were flattened into a 1 dimensional vector 
of size $144$. Three fully-connected layers of output sizes $144, 144, 12$ were used, each followed by ReLU and dropout with $p=0.25$,
after which a final fully connected layer with output size $N_k = 6$ was used. 
The treatment indicator was concatenated to these activations for the final prediction.
We used a batch size of 40 and the Adam optimizer \cite{adam} with a learning rate of 0.001 and no weight-decay. 

\subsubsection{Results}
We calculated 3 baseline models for comparison: (1) ignoring all image information and using only the
treatment indicator, or linear regression on the 'oracle' data $\{t, x, z, y\}$ with (2) and 
without (3) conditioning on the collider $x$. 
Through the sampling scheme, along with ambiguity in manual nodule segmentations and limitations 
of statistical learning from finite data, there is inherent prediction error for $y$ and $x$.
We estimated the $MSE$ of this inherent error by predicting the ground truth labels $x$ and $z$ with 
a separate run of the same CNN architecture by replacing $y$ with $z$. 
For fair comparison of the methods, in the regression baseline models we replaced $x, z$ 
by $x', z'$ by adding gaussian noise to the simulated $x, z$ based on the
$MSE$ of the ground truth run for both variables.
We compare the 'curve fitting' approach of conditioning on the entire image 
for predicting $y$ (BiasNet) with the proposed method (CausalNet). As presented 
in Table \ref{tabresults}, the proposed method perfectly separates the biasing 
effect of the collider $x$ on the estimated treatment effect, and attains a prediction 
error very close to the ideal expected loss for predicting $y$.

\begin{table}[]
  \centering
  \begin{tabular}{llll}
  \textbf{model} &  \textbf{variables}    & \textbf{$MSE_y$} & \textbf{$ATE$} \\ \hline
  Regression     & $t$                    & 2.74             & 1.01           \\ %
  Regression     & $t, x', z'$            & 1.35             & 0.42           \\ %
  Regression*    & $t, z'$                & 2.07             & 1.01           \\ 
  BiasedNet      & $t, image$               & 1.72             & 0.41           \\ %
  CausalNet      & $t, a_j^k (j>1)$       & \textbf{2.00}    & \textbf{0.99}   \\ \hline
  \end{tabular}
  \caption{Results: Mean squared error for survival ($MSE_y$) along with 
          estimated Average Treatment Effect ($ATE$). 
          The linear regression metrics are the expected outcomes according to 
          whether or not the model conditions on the collider $x$.
          Regression* is the optimal value for our setup:
          (1) predicting the outcome based on relevant prognostic information from the image while 
          (2) retaining a valid estimate of the treatment effect.
          All metrics were calculated on the validation set}
  \label{tabresults}
  \end{table}
\subsection{Discussion}

We provide a realistic medical example where plain \textit{curve fitting} with deep learning
will lead to biased predictions that do not generalize to the setting where we \textit{intervene} on treatment. 
By utilizing prior knowledge about the world in the design of the CNN-architecture and optimization scheme,
accurate survival predictions were feasible with an unbiased estimate of the treatment effect. 
Our experiments demonstrate that deep learning can in principle be combined with insights from causal inference. 
Possible directions for extension of our experiments are introducing different data 
generating mechanisms, for example with a treatment effect modifier or with statistical dependence between factors of variation within the image.
In addition, similar approaches can be explored for medical images from different sources (e.g. pathology slides), 
or different data domains such as audio or natural language. We leave these extensions for further work.

To attain the goal of personalized treatment recommendations with artificial intelligence,
methods combining machine learning with causal inference need to be further developed. 
Our experiments provide an example of how deep learning and structural causal models can be combined
and are a small step forward towards personalized health care.

\subsection{Acknowledgements}
We kindly thank Pim de Jong, Tim Leiner and Joost Verhoeff.
Furthermore, we thank NVIDIA for supplying us with a Quadro P-6000 GPU through the academic seeding grant program, which was used in the experiments.

\printbibliography

\end{document}